\documentclass[sigconf]{acmart}

\usepackage{booktabs} 
\usepackage{amsmath}
\usepackage{subcaption}

\copyrightyear{2019} 
\acmYear{2019} 
\setcopyright{acmcopyright}
\acmConference[GECCO '19]{Genetic and Evolutionary Computation Conference}{July 13--17, 2019}{Prague, Czech Republic}
\acmBooktitle{Genetic and Evolutionary Computation Conference (GECCO '19), July 13--17, 2019, Prague, Czech Republic}
\acmPrice{15.00}
\acmDOI{10.1145/3321707.3321849}
\acmISBN{978-1-4503-6111-8/19/07}

\begin{document}

\title{Intentional Computational Level Design}


\author{Ahmed Khalifa}
\affiliation{%
  \institution{New York University}
  \streetaddress{5 Metrotech Center}
  \city{Brooklyn}
  \state{New York}
  \postcode{11201}
}
\email{ahmed@akhalifa.com}

\author{Michael Cerny Green}
\affiliation{%
  \institution{New York University}
  \streetaddress{5 Metrotech Center}
  \city{Brooklyn}
  \state{New York}
  \postcode{11201}
}
\email{mcgreentn@gmail.com}

\author{Gabriella Barros}
\affiliation{%
  \institution{modl.ai}
  \streetaddress{}
  \city{Macei\'{o}} 
  \country{Brazil}
}
\email{gabbbarros@gmail.com}

\author{Julian Togelius}
\affiliation{%
  \institution{New York University}
  \streetaddress{5 Metrotech Center}
  \city{Brooklyn}
  \state{New York}
  \postcode{11201}
}
\email{julian@togelius.com}

\renewcommand{\shortauthors}{Khalifa et al.}

\begin{abstract}
The procedural generation of levels and content in video games is a challenging AI problem. Often such generation relies on an intelligent way of evaluating the content being generated so that constraints are satisfied and/or objectives maximized. In this work, we address the problem of creating levels that are not only playable but also revolve around specific mechanics in the game. We use constrained evolutionary algorithms and quality-diversity algorithms to generate small sections of Super Mario Bros levels called \textit{scenes}, using three different simulation approaches: \textit{Limited Agents}, \textit{Punishing Model}, and \textit{Mechanics Dimensions}. All three approaches are able to create scenes that give opportunity for a player to encounter or use targeted mechanics with different properties. We conclude by discussing the advantages and disadvantages of each approach and compare them to each other.
\end{abstract}

%
%



\maketitle

\section{Introduction}

Level generation is a core problem within game AI. Game levels are areas that can or need to be traversed in order to finish the game. 
These levels are constructed from a set of building blocks defined by the game, such as doors, walls, and NPCs of various kinds. 

Game levels need to fulfill various specifications; in most games, they need to be beatable as well as fun and enjoyable to play. They can contribute to the plot of the story the game is trying to tell. But most crucially, levels can and should challenge the player by introducing or revisiting various mechanics that the player encounters in the game. By keeping the player on their toes, the levels constantly challenge the player, forcing them to adapt or learn different strategies and mechanics to win. Building and evaluating levels that focus on core mechanics is complicated, and automating the process is largely an unsolved problem.

We previously proposed that levels or maps that teach mechanics could be generated in order to provide tailored or educational experiences~\cite{green2017press,green2018atdelfi,green2018generating}.
However, targeting specific level features or mechanics during level generation is a difficult AI problem, and few examples of this can be found. One of the reasons for this is in the evaluation of the level itself. How can the generator automatically verify that the level meets these requirements?

In this paper, we introduce three methods which generate and evaluate mechanic-centric levels. Some of the methods are faster while others capture a wider range of mechanics, but all of them generate levels which emphasize opportunities for the player to experience the mechanic that was targeted during level generation.

\section{Background}

\subsection{Mario AI Framework}

\emph{Infinite Mario Bros.} (IMB) was developed by Markus Persson~\cite{persson2008infinite} as a public domain clone of \emph{Super Mario Bros}(Nintendo 1985). Much like the original, IMB consists of Mario (the player's avatar) moving horizontally on a two-dimensional level towards a goal. Mario can be in one of three possible states: small, big and fire. Each state increases the amount of times Mario can take damage without failing the level and also give Mario special abilities. Mario can move left and right, jump, run, and, when in the fire state, shoot fireballs. The player returns to the previous state if they take damage and dies when taking damage if in the small state. They can also die when falling down a gap in any state. Additionally, unlike the original game, IMB allows for automatic generation of levels.

The Mario AI framework is a popular benchmark for research on artificial intelligence built on top of IMB~\cite{karakovskiy2012mario}, having been used in AI competitions in the past~\cite{karakovskiy2012mario,togelius2013mario}. It improved on limitations of IMB's level generator, and several techniques have been applied to automatically play~\cite{karakovskiy2012mario} or create levels~\cite{shaker20112010,sorenson2010towards}, some of which are search-based methods.

\subsection{Search Based Procedural Content Generation}

Search-based PCG is a technique that explores possible generative space via search methods. One example is evolutionary algorithmic search~\cite{togelius2011search}, a type of optimization inspired by Darwinian evolutionary concepts like reproduction, fitness, and mutation. Search-based methods can be used to generate levels and content within games. Ashlock et al.~\cite{ashlock2010automatic,ashlock2011search} developed a variety of ways to do this, such as optimized puzzle generation for different difficulties, and stylized cellular automata evolution for cave generation~\cite{ashlock2015evolvable}. In a different approach, McGuinness et al.~\cite{mcguinness2011decomposing} created a system that decomposed level generation into an evolutionary micro-macro process. 

We have previously used evolution to generate levels in several domains, such as General Video Game AI and PuzzleScript~\cite{khalifa2015automatic}. Later work by Khalifa et al.~\cite{khalifa2018talakat} used constrained Map-Elites, a hybrid evolutionary search, to generate levels for bullet hell games, and, using automated playing agents with different parameters to mimic various human play-styles, \textit{Talakat} was able to target multiple types of levels simultaneously. Shaker et al.~\cite{shaker2013evolving} evolved levels for \emph{Cut the Rope} (ZeptoLab 2010) using constraint evolutionary search where the fitness measures the playability of a level using playable agents. Khalifa et al.~\cite{khalifa2015literature} offers a literature review of search based level generation within puzzle games. Finally, Green et al. developed an evolutionary level generator which curated training data for a deep reinforcement learning agent~\cite{green2019evolutionarily}, procedurally evolving levels to continually challenge it in training.

Very few attempts have been made to generate game levels that specifically enable or require particular mechanics. The educational game \textit{Refraction} (Center for Game Science at the University of Washington 2010) generates levels using answer set programming to target particular level features~\cite{smith2012case}. We previously proposed a method to automatically generate these experiences using search-based procedural content generation methods, which the current work directly builds on and extends significantly~\cite{green2018generating}.

\subsection{Level Generation for the Mario AI Framework}
The Mario AI framework's organizers hosted competitions for level generation in 2010 and 2012~\cite{togelius2013mario}. Horne et al.~\cite{horn2014comparative} assembled a comprehensive list of these generators, briefly described here, in addition to other generators written outside the competition. 

The \textit{Notch} and the \textit{Parameterized-Notch} generators create levels from left to right by adding game elements through probability and performing basic checks to ensure playability~\cite{shaker2011feature}. Similarly, the \textit{Hopper} generator also designs levels from left to right and adds game elements through probability, but unlike the previous generators, these probabilities adapt to player performance which results in a dynamic level generator~\cite{shaker20112010}. \textit{Launchpad} is a rhythm-based level generator which uses grammars to create levels within rhythmical constraints~\cite{smith2011launchpad}. The \textit{Occupancy-Regulated Extension} generator glues small hand-authored chunks together into complete levels~\cite{shaker20112010}. The \textit{Pattern-based} generator uses \emph{slices} taken from the original \emph{Super Mario Bros} (Nintendo 1985) to evolve levels, where the fitness function counts the number of occurrences of specified sections of slices, with the objective of maximizing the number of different slices~\cite{dahlskog2013patterns}. The \textit{Grammatical Evolution} generator evolves levels with design grammars. Levels are represented as instructions for expanding design patterns, and the fitness function measures the number of items in the level and the number of conflicts between the placement of these items. In previous work, we~\cite{green2018generating} generated small level ``scenes'' that try to teach the player specific mechanics, like high jumping or stomping on enemies. 

\section{Methods}
\begin{figure}[t]
    \centering
    \includegraphics[width=0.5\linewidth]{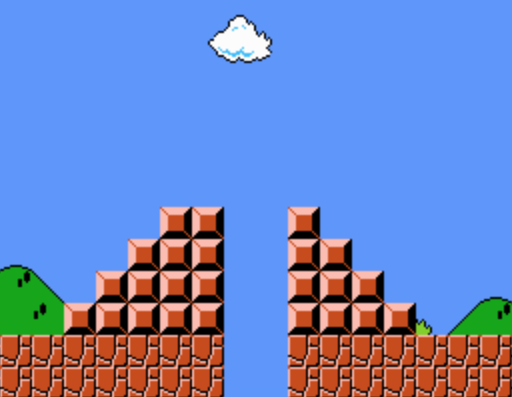}
    \vspace{-5pt}
    \caption{Super Mario Bros scene where Mario need to jump over a gap from the first pyramid to the second pyramid.}
    \label{fig:smb_scene}
    \vspace{-10pt}
\end{figure}

In this project, we generate scenes for the Mario AI Framework. A \textit{scene} is a small section of the level that encapsulate a certain idea~\cite{anthropy2014game}. For example, figure~\ref{fig:smb_scene} shows a scene from the first Super Mario Bros (Nintendo 1985) where the idea is to make sure the player learns about the jump mechanic by accurately jumping from the first pyramid to the second pyramid without falling in the gap. We explore three different search-based approaches to generate scenes: ``Limited Agents'', ``Punishing Model'' and ``Mechanics Dimensions''.

The ``Limited Agents'' technique uses two separate agents to generate levels. One of these agents is, ideally, a ``perfect'' agent, meaning that it is always able to find a way to beat the level if any such way exists. The second agent is a variant of a perfect agent but limited in some mechanic-based way (for example, being completely unable to jump in the game). The technique strives to make sure that, given a generated scene, a perfect agent (i.e. an agent that knows all mechanics) can beat it, but an agent that lacks a required mechanic cannot~\cite{green2018generating}.
The ``Punishing Model'' uses one agent (the perfect agent) but with two different forward models.
One of these forward models is perfect while the other punishes the agent for causing certain mechanics during play.

These two techniques use the \emph{Feasible Infeasible 2-Population (FI2Pop)} algorithm~\cite{kimbrough2008feasible} to generate levels, which will be discussed in details in the next subsection. The last approach, ``Mechanics Dimensions'', uses the \emph{Constrained Map-Elites (CMElites)} algorithm~\cite{khalifa2018talakat} to generate passable scenes using the perfect agent and recording every mechanic that fires during that playthrough. The algorithm places the chromosome in the appropriate cell in the CMElites' map depending on the fired game mechanics.

\subsection{Evolutionary Algorithms}
This project uses two different types of evolutionary search to generate scenes: \emph {Feasible-Infeasible 2-Population (FI2Pop)} and \emph{Constrained Map-Elites}. As mentioned above, the ``Punishing Model'' and the ``Limited Agents'' methods used FI2Pop, while the ``Mechanics Dimensions'' approach used Constrained Map-Elites. The following subsections describe these in more detail.

\subsubsection{Feasible Infeasible 2-Population}
FI2Pop is a genetic algorithm that uses a dual-population technique~\cite{kimbrough2008feasible}. One of these, the ``feasible'' population, attempts to improve the overall quality of chromosomes contained within. These chromosomes can occasionally break constraints while improving and are consequently placed into the ``infeasible'' population. The fitness function of the infeasible population drives chromosomes toward parameters that bring them back within the constraints of the ``feasible'' population.

\subsubsection{Constrained Map-Elites}
Constrained Map-Elites~\cite{khalifa2018talakat} is a hybrid genetic algorithm that uses FI2Pop on top of the Map-Elites algorithm~\cite{mouret2015illuminating}. As with standard Map-Elites, it maintains a map of $n$-dimensions instead of using a population. Each dimension is sampled into more than one value
which divides the map into a group of cells. The dimensions usually correspond to a chromosome behavior such as the number of coins collected in a Super Mario Bros playthrough. However, where in the standard Map-Elites each cell stores one chromosome or a population, in Constrained Map-Elites each cell stores two populations similar to FI2Pop: one for the feasible chromosomes and the other for the infeasible chromosomes.
Chromosomes can be moved between cells (if they change their dimensions) and/or between populations within their cell (if they fit within the feasible population's constraints or not).

\subsection{Scene Representation}
\begin{figure}[t]
    \centering
    \includegraphics[width=0.8\linewidth]{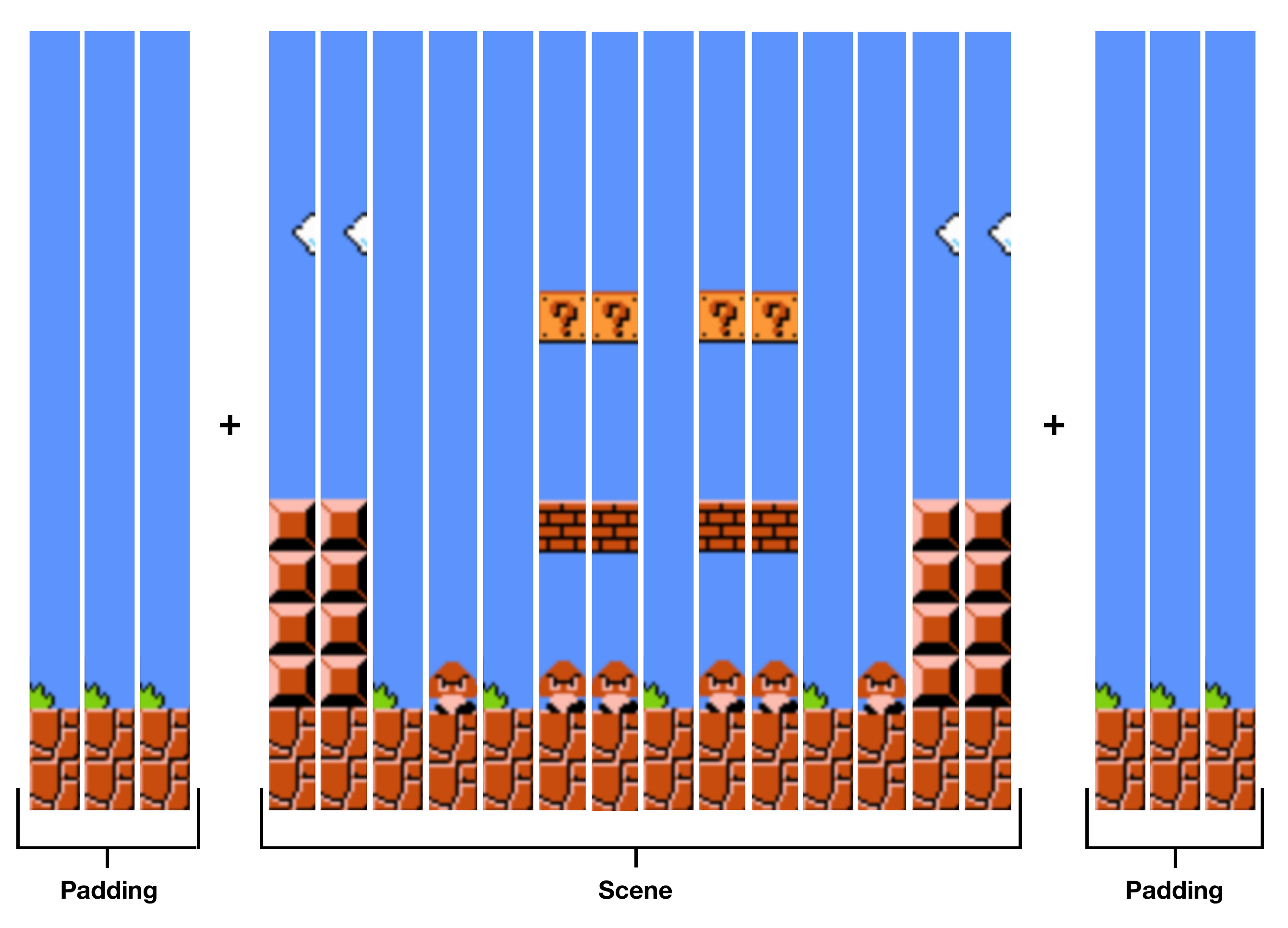}
    \vspace{-5pt}
    \caption{The chromosome consists of 14 vertical slices padded with 3 floor slices on both ends.}
    \label{fig:smb_slices}
    \vspace{-10pt}
\end{figure}

Chromosomes are represented the same way in all three approaches. We assumed that a \emph{Super Mario Bros} (Nintendo 1985) screen is equivalent to a scene. Therefore, chromosomes are represented as a sequence of vertical slices sampled from the first \emph{Super Mario Bros}, similar to the representation used by Dahlskog and Togelius~\cite{dahlskog2013patterns}. A scene consists of 14 vertical slices padded with three floor slices surrounding the scene, as shown in Figure~\ref{fig:smb_slices}. This padding is necessary to ensure that there is a safe area where Mario starts and finishes the level. Additionally, these slices were collected from the levels presented in the Video Game Level Corpus (VGLC)~\cite{summerville2016vglc}. There are $3721$ vertical slices in the corpus with $528$ unique vertical slices. Slices are sampled based on their percentage of appearance in the VGLC. For example: a flat ground slice will have a higher chance to be picked compared to other slices.

\subsection{Scene Evaluation}
Scenes are evaluated in relation to specific constraints, which differ slightly among the different techniques and will be discussed in later sections. If the scene satisfies these constraints, it is also evaluated in terms of simplicity. The simplicity fitness is unified between all the techniques and tries to reduce the entropy in the scene. The entropy calculation makes sure that generated scenes place tile efficiently and have high horizontal consistency. Equation~\ref{eq:fitness} shows the fitness calculation equation.

\begin{equation}\label{eq:fitness}
    fitness = \left( 0.2 \cdot \left( 1 - H\left(X\right)\right)\right) + \left(0.8 \cdot \left(1 - H\left(\bar{X}\right)\right)\right)
\end{equation}

where $X$ is the set of different used tile types in the scene and $\bar{X}$ is the set of horizontal tile changes in the scene. We weighted the latter (i.e. the horizontal change entropy part) higher than the former, as the abrupt changes in tiles occur more often when there are many different types of objects, making the scene look noisy. For example, if there are two scenes with the same amount of objects, they will have the same entropy, but the order of their arrangement will affect the horizontal change entropy calculations. 

Regarding the aforementioned constraints, all approaches make sure that the scene is playable by using Robin Baumgarten's A* agen, the winner of the first Mario AI competition~\cite{togelius20102009}. Besides the playability constraint, another constraint is included to make sure that the scenes contain the targeted mechanic, whatever that may be. This additional constraint will be discussed in detail for each approach in the following subsections.

\subsubsection{Limited Agents}
For this technique, we used two agents: Robin Baumgarten's A* agent (the ``perfect agent'') and a modified version of it that we called the ``limited agent''. It is important to note here that the A* is not, in reality, perfect. However, it is about as close to perfect an agent for Mario as possible at this current moment. Similar to previous work~\cite{green2018generating} during the scene evaluation phase, the constraint is satisfied when the perfect agent can win the scene while the limited agent cannot.

In case the constraint is not satisfied, two scenarios may occur. If both agents win the level, the fitness of the chromosome is 0. If neither agent wins the level, then the chromosome's fitness is proportional to the difference of the distance traveled by the two agents.
Maximizing the relative distance between the perfect agent and the limited agent is directly proportional to satisfying the constraint. A good scene can be defined as the perfect agent traversing the entire scene while the limited agent travels less. Equation~\ref{eq:la_constraints} shows the constraints equation:

\begin{equation}\label{eq:la_constraints}
    constraints = \begin{cases}
        1 & \text{if $A_{perf} = 1$ \& $A_{limit} = 0$}\\
        \frac{d_{perf} - d_{limit}}{d_{scene}} & \text{otherwise}
    \end{cases}
\end{equation}

where $A_{perf}$ and $A_{limit}$ are the result of the perfect agent and the limited agent playing the scene, respectively being $1$ when the agent wins and $0$ otherwise; $d_{perf}$ and $d_{limit}$ are the distances traveled by the perfect- and the limited agent. $d_{scene}$ is the length of the full scene, used to normalize the result to the $-1$ and $1$ range.

\begin{table}[t]
    \centering
    \begin{tabular}{|p{0.25\linewidth}|p{0.65\linewidth}|}
        \hline
        Agent & Limitation \\
        \hline
        \hline
        No Run & can't press the run button.\\
        \hline
        Limited Jump & can't hold the jump button for too long.\\
        \hline
        Enemy Blind & doesn't see enemies.\\
        \hline
    \end{tabular}
    \caption{Agents' Limitation}
    \label{tab:la_types}
    \vspace{-15pt}
\end{table}

In this work, we used three different types of limited agents, shown in table~\ref{tab:la_types}. We limited ourselves to the same agents from previous work~\cite{green2018generating} because adding additional mechanic based limitations has less impacting for this technique. For example, in the case of the coin collection mechanic, an agent that is coin-blind won't die or stop the game's execution if it didn't collect a coin.

\subsubsection{Punishing Model}
This technique uses only one agent (the perfect agent), but relies on two different forward models. The first forward model, called ``normal model'', behaves as expected in the game. However, the second forward model, the ``punishing model'', makes the agent believe that it will die when a certain mechanic is fired. For example, the agent believes it will die if it stomps on an enemy and consequently tries to avoid doing so. The constraints' value is calculated using equation~\ref{eq:la_constraints}, using the punishing forward model's results instead of the limited agent's.

\begin{table}[t]
    \centering
    \begin{tabular}{|p{0.25\linewidth}|p{0.65\linewidth}|}
        \hline
        Mechanic Model & Punishment \\
        \hline
        \hline
        High Jump & kills the player if they hold the jump button for too long.\\
        \hline
        Speed & kills the player if they exceed a certain $x$ velocity.\\
        \hline
        Stomp & kills the player if they stomp an enemy.\\
        \hline
        Shell Kill & kills the player if they kill an enemy using a koopa shell.\\
        \hline
        Mushroom & kills the player if they collect a mushroom.\\
        \hline
        Coin & kills the player if they collect a coin.\\
        \hline
    \end{tabular}
    \caption{Mechanic models and their perceived punishments}
    \label{tab:pm_types}
    \vspace{-15pt}
\end{table}

Table~\ref{tab:pm_types} shows six different punishing models used in this project. These are six different basic mechanics that appear in the original \emph{Super Mario Bros}. We exclude mechanics that only appeared in following titles, such as wall jumping and ground pound.

\subsubsection{Mechanics Dimensions}
From an architectural standpoint this is simplest of the three techniques, using only one agent (the perfect agent) and one forward model (the normal forward model). The constraints value is equal to $1$ if the agent can beat the scene, and directly proportional to the traversed distance otherwise, as shown in Equation~\ref{eq:mc_constraints}.

\begin{equation}\label{eq:mc_constraints}
    constraints = \begin{cases}
    1 & \text{if $A_{perf} = 1$}\\
    \frac{d_{perf}}{d_{scene}} & \text{otherwise}
    \end{cases}
\end{equation}

where $A_{perf}$ is the result of a perfect agent playing the scene, $d_{perf}$ is the distance traversed by the agent, and $d_{scene}$ is the scene length, used to normalized the result to $0 -- 1$.

\begin{table}[t]
    \centering
    \begin{tabular}{|p{0.25\linewidth}|p{0.65\linewidth}|}
        \hline
        Dimension & Description \\
        \hline
        \hline
        Jump & is $1$ if the player jumped in the level and $0$ otherwise.\\
        \hline
        High Jump & is $1$ if the player jumped higher than a certain value and $0$ otherwise.\\
        \hline
        Long Jump & is $1$ if the player's horizontal traversed distance after landing is larger than a certain value and $0$ otherwise.\\
        \hline
        Stomp & is $1$ if the player stomped on an enemy and $0$ otherwise.\\
        \hline
        Shell Kill & is $1$ if the player killed an enemy using a koopa shell and $0$ otherwise.\\
        \hline
        Fall Kill & is $1$ if an enemy dies because of falling out of the scene and $0$ otherwise.\\
        \hline
        Mushroom & is $1$ if the player collected a mushroom during the scene and $0$ otherwise.\\
        \hline
        Coin & is $1$ if the player collected a coin during the scene and $0$ otherwise.\\
        \hline
    \end{tabular}
    \caption{Constrained Map-Elites' dimensions.}
    \label{tab:mc_types}
    \vspace{-15pt}
\end{table}

After calculating the constraints, the playthrough is analyzed to extract the different types of mechanics that were fired during play. This extracted information is used to create binary dimensions for the constrained map-elites algorithm. Table~\ref{tab:mc_types} shows the different mechanics extracted from the playthrough. These mechanics were selected in order to be similar to the six mechanics used in the punishing model approach, but also include additional mechanics like ``Jump'' and ``Fall Kill,'' which were more difficult or impossible to represent/isolate in the punishing model or limited agent approach.
For example, fall kill was added to help differentiate between scenes that requires overcoming enemies through action and scenes where enemies will fall out of the scene if the player has waited or even played imperfectly (killing through inaction). Furthermore, the different jump types were added to differentiate between scenes that required (1) running and jumping, or (2) holding the jump button for a bit longer to high jump, or (3) just a small tap.

\section{Results}
We ran all three approaches for $1000$ generations (for FI2Pop) / iterations (for Constrained Map-Elites), where each generation/iteration consists of $100$ new chromosomes. We ran each experiment with its different configurations for $5$ times. In the Constrained Map-Elites, infeasible population size was fixed to twenty chromosomes, and the feasible population size was fixed to one chromosome. These values were selected based on our current available resources and also to make sure that generation algorithm will finish in under than 12 hours using 25 CPU cores.

We used a two point crossover that swaps any number of slices, ranging from a single slice to the full scene. The mutation replaces a single slice with a random slice sampled from slices in the original \emph{Super Mario Bros}. For FI2Pop, elitism was used to preserve the best chromosome across generations. Robin Baumgarten A* agent is not an actual perfect agent, rather, but only a close approximation of one. As a result, chromosomes do not always receive the same constrained value over multiple playthroughs (the agent may perform different actions in the exact same scenario). To counter this, we keep the lowest constrained value after evaluation of the chromosomes between generations.


The following subsections present the results of each of our three approaches. These results were evaluated by members of our team through playing the generated levels themselves and capturing their observations.

\subsection{Limited Agents}

\begin{figure}
    \centering
    \includegraphics[width=0.8\linewidth]{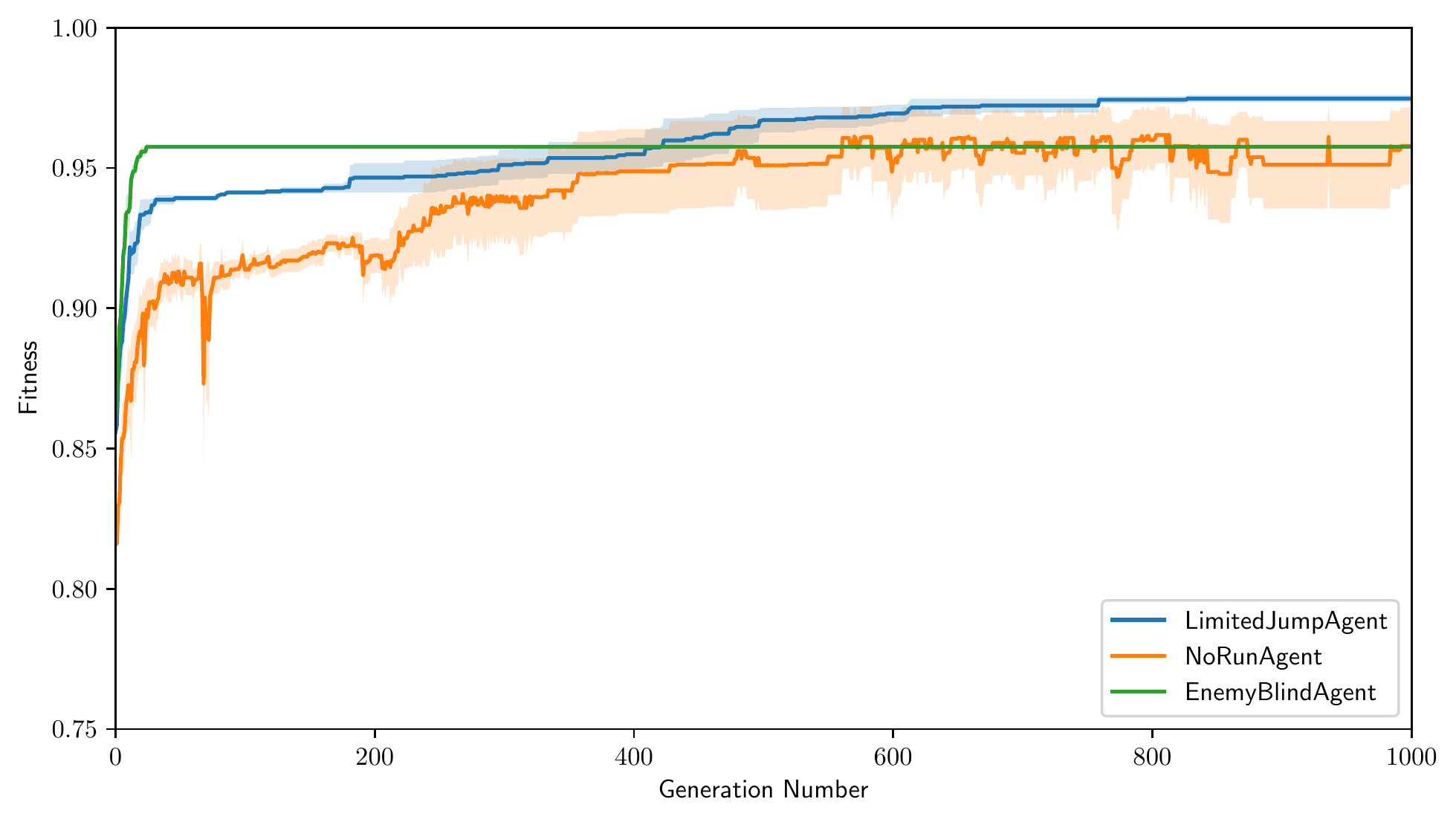}
    \caption{The average maximum fitness value and the standard error of the limited agents approaches over the $5$ runs.}
    \vspace{-5pt}
    \label{fig:la_graph}
    \vspace{-10pt}
\end{figure}

The ``Limited Agents'' approach finds the targeted technique faster than any other approach. Figure~\ref{fig:la_graph} shows the average maximum fitness for each of the targeted mechanics. The \textit{Enemy Blind} agent was the fastest, finding the best chromosome by generation eight. Its constraints were simpler than that of other agents, as it only cared about overcoming enemies instead of more specific mechanics like stomping on them. On the other hand, the \textit{No Run} agent quickly found a good, albeit unstable, solution, resulting in a noisy fitness line. Unstable solutions are solutions where constraints are satisfied due to the agent being imperfect, thus their constraints' values change whenever a better playthrough is found, causing them to sometimes move to the infeasible population instead.

\begin{figure*}
    \centering
    \begin{subfigure}[t]{.28\linewidth}
        \centering
        \includegraphics[width=0.95\linewidth]{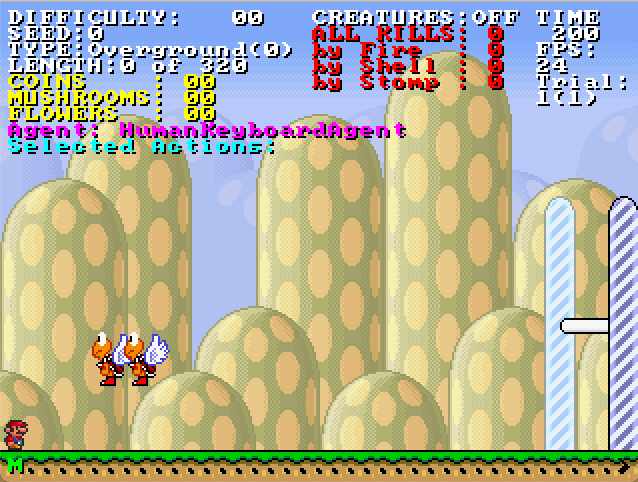}
        \caption{No Run Agent}
        \label{fig:la_a}
    \end{subfigure}
    \begin{subfigure}[t]{.28\linewidth}
        \centering
        \includegraphics[width=0.95\linewidth]{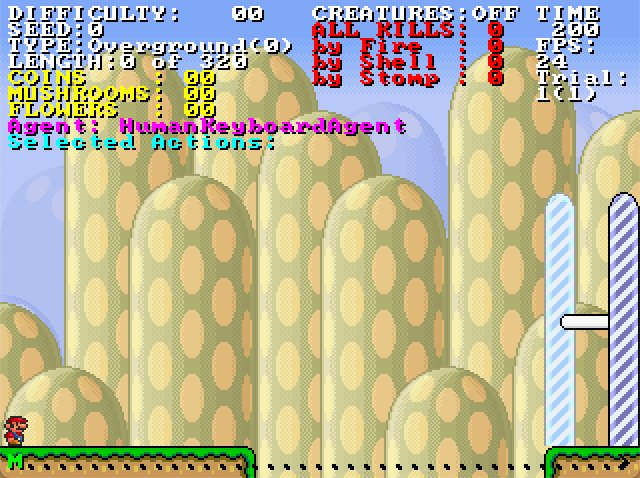}
        \caption{Limited Jump Agent}
        \label{fig:la_b}
    \end{subfigure}
    \begin{subfigure}[t]{.28\linewidth}
        \centering
        \includegraphics[width=0.95\linewidth]{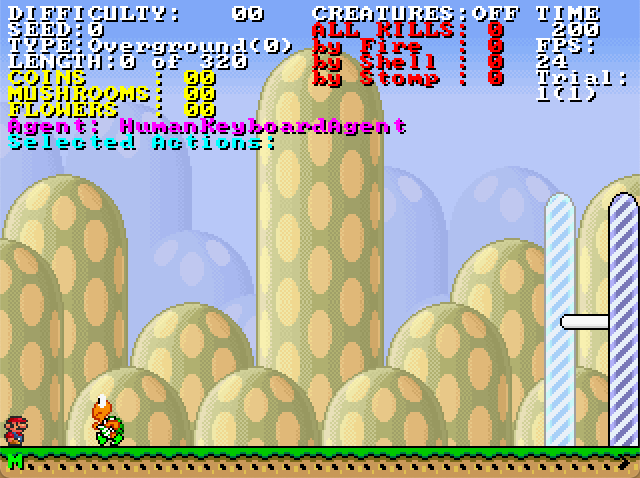}
        \caption{Enemy Blind Agent}
        \label{fig:la_c}
    \end{subfigure}
    \vspace{-5pt}
    \caption{The results for the limited agent approach after 1000 generations.}
    \label{fig:la_results}
    \vspace{-10pt}
\end{figure*}

Figure~\ref{fig:la_results} shows the best chromosome at the end of the $1000$ generations for each targeted mechanic from one of the $5$ runs. Using both \textit{Enemy Blind} and \textit{Limited Jump}, the generated scenes satisfy the targeted mechanics. The \textit{Enemy Blind} agent's scene required the agent to overcome a koopa troopa in order to proceed, either by stomping on it or jumping over it. The \textit{Limited Jump} agent's scene requires both running and holding the jump button for many time ticks to pass a gap. However, the \textit{No Run} agent's scene can be passed without running if the player waits (until the koopa paratroopa lands) and then move to finish the level, or by jumping on top of the koopa paratroopas. This scene was generated due to one of the imperfect playthroughs, where the agent moves forward immediately when the scene starts, dying as soon as the koopa paratroopa lands on them, rendering that scene unstable. Also of note, the \textit{Limited Jump} agent's scene can act as a \textit{No Run} agent's scene as well, as the gap cannot be passed without running and holding the jump button. From analyzing the best chromosome from all the $5$ runs, we found that all the generated levels are different from each other except for the \textit{Enemy Blind Agent} experiment. As all the generated levels looks similar to figure~\ref{fig:la_c} but with different enemy x positions or types.

\subsection{Punishing Model}

\begin{figure}
    \centering
    \includegraphics[width=0.8\linewidth]{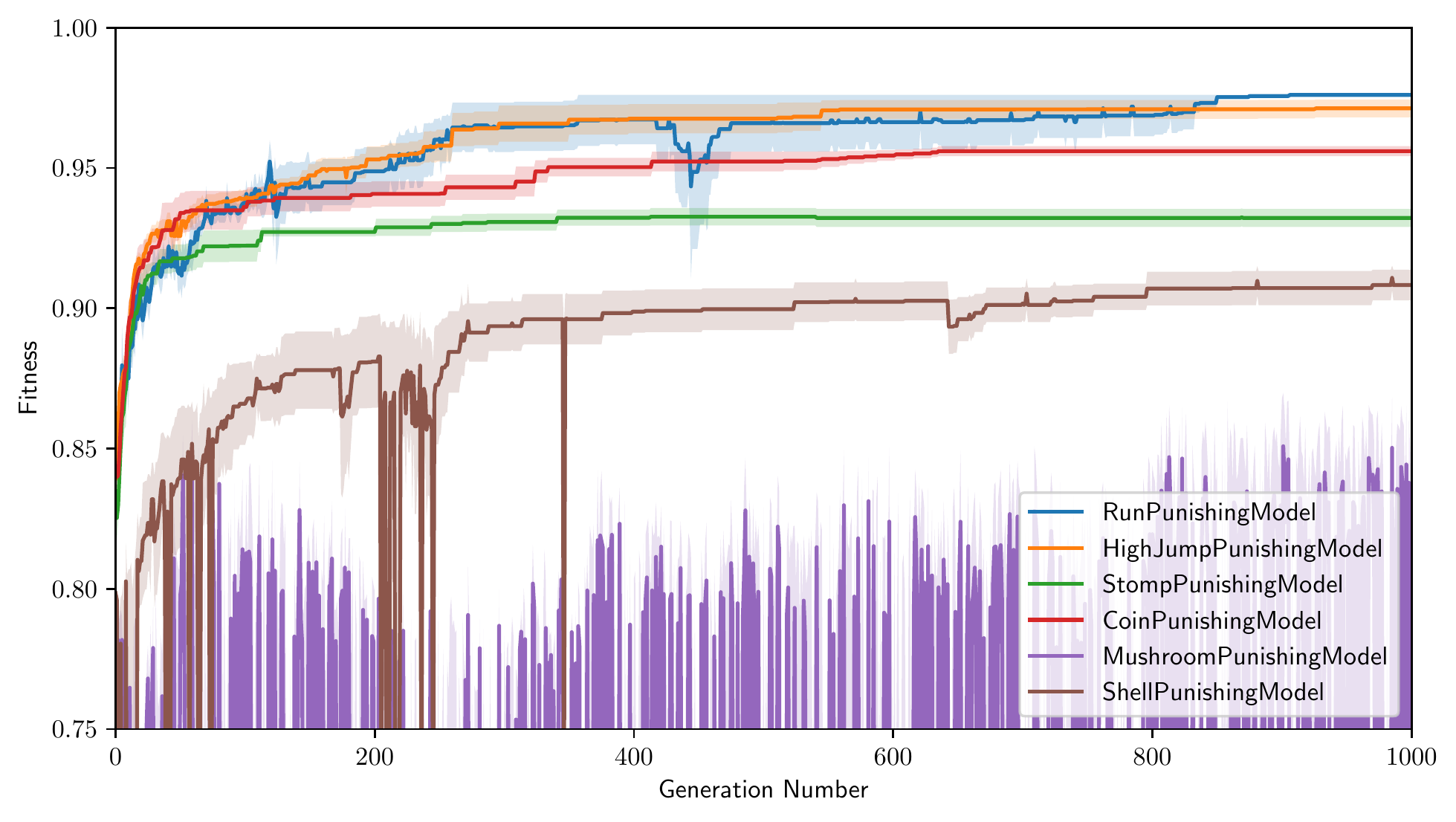}
    \vspace{-5pt}
    \caption{The average maximum fitness value and the standard error of the punishing model approaches over $5$ runs.}
    \label{fig:pm_graph}
    \vspace{-10pt}
\end{figure}

The punishing model is the slowest technique to find a targeted mechanic as the A* agent spent every tick trying to compute alternative paths to not die from the punishing forward model. Figure~\ref{fig:pm_graph} shows the average maximum fitness for each of these mechanics. Results show several unstable scenes for most iterations, before stabilizing, except for the \textit{Mushroom Punishing Model}, which never stabilized and always fluctuated between a $0$ fitness (no scenes satisfied the constraints) and above a $0.8$. This instability was caused from forcing the AI to try find every possible path at every tick. Occasionally the agent missed a viable solution without triggering the mechanic in one generation but later found it in the next, therefore changing the chromosome's constraints and fitness.

\begin{figure*}
    \centering
    \begin{subfigure}[t]{.28\linewidth}
        \centering
        \includegraphics[width=0.95\linewidth]{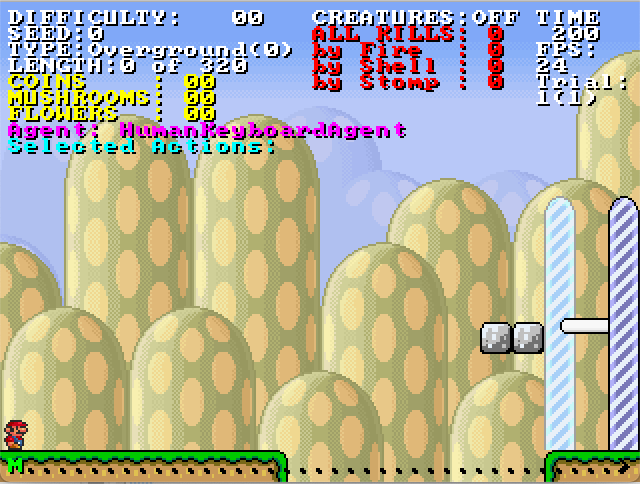}
        \caption{High Jump Punishing Model}
        \label{fig:pm_a}
    \end{subfigure}
    \begin{subfigure}[t]{.28\linewidth}
        \centering
        \includegraphics[width=0.95\linewidth]{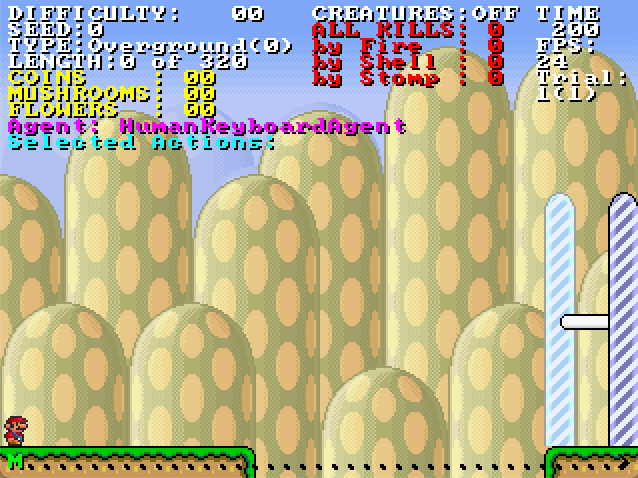}
        \caption{Speed Punishing Model}
        \label{fig:pm_b}
    \end{subfigure}
    \begin{subfigure}[t]{.28\linewidth}
        \centering
        \includegraphics[width=0.95\linewidth]{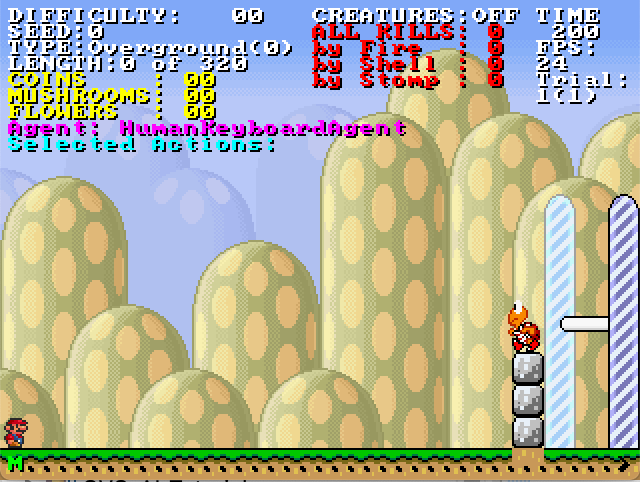}
        \caption{Stomp Punishing Model}
        \label{fig:pm_c}
    \end{subfigure}
    \begin{subfigure}[t]{.28\linewidth}
        \centering
        \includegraphics[width=0.95\linewidth]{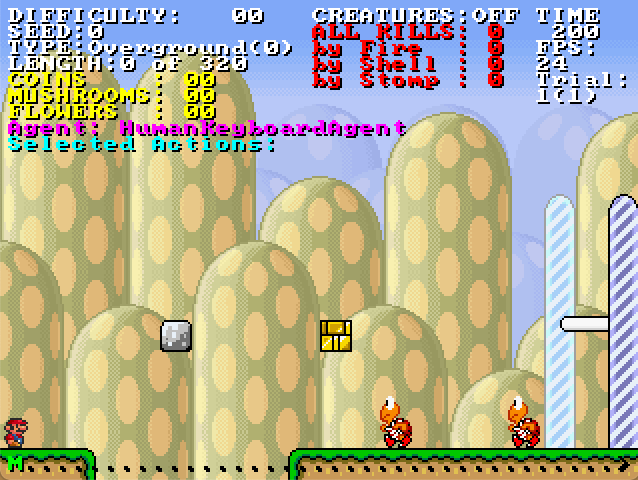}
        \caption{Shell Kill Punishing Model}
        \label{fig:pm_d}
    \end{subfigure}
    \begin{subfigure}[t]{.28\linewidth}
        \centering
        \includegraphics[width=0.95\linewidth]{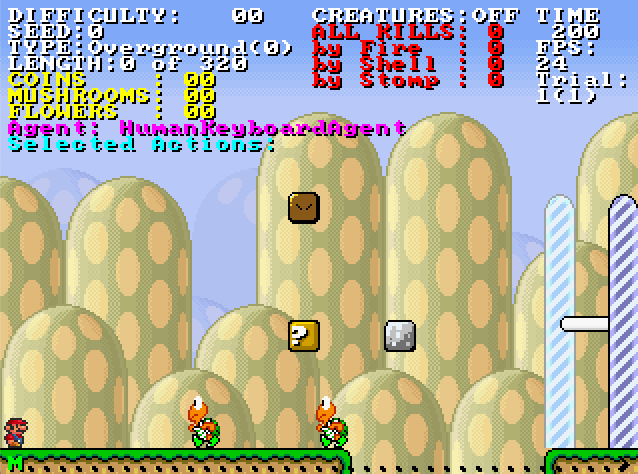}
        \caption{Mushroom Punishing Model}
        \label{fig:pm_e}
    \end{subfigure}
    \begin{subfigure}[t]{.28\linewidth}
        \centering
        \includegraphics[width=0.95\linewidth]{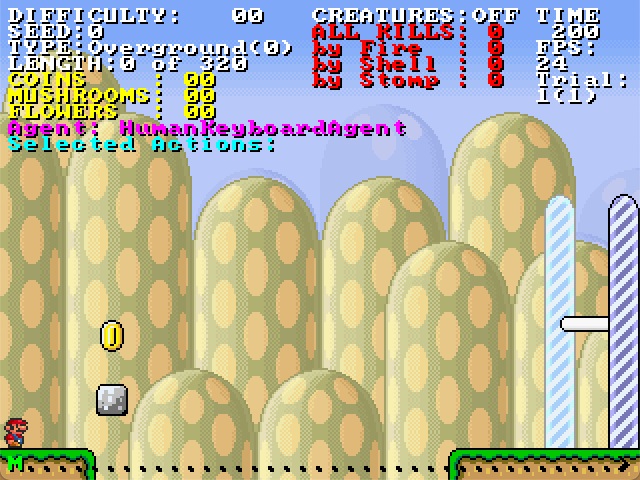}
        \caption{Coin Punishing Model}
        \label{fig:pm_f}
    \end{subfigure}
    \vspace{-5pt}
    \caption{The results for the punishing model approach after 1000 generations.}
    \label{fig:pm_results}
    \vspace{-10pt}
\end{figure*}

Figure~\ref{fig:pm_results} shows results for each targeted mechanic from one of the $5$ runs. The scenes generated for the \textit{High Jump-}, \textit{Speed-}, \textit{Stomp-}, and \textit{Coin Punishing Model} are impossible to win unless the targeted mechanic is used. The remaining two generated scenes rely heavily on the use of the targeted mechanic, but a creative agent (or human) may notice other solutions that do not involve that mechanic. 
The existence of these scenes occur because the agent is imperfect, therefore it cannot exhaustively search for all possible solutions to ensure they require use of the mechanic. By looking at all the best chromosomes from all the $5$ runs, we found that all the generated scenes, while not explicitly requiring the mechanic, do provide all the \textit{ingredients} needed for the mechanic to occur. For example, the shell kill punishing model requires having at least two koopas, speed punishing model requires having a large gap, etc.

\subsection{Mechanics Dimensions}

\begin{figure}
    \centering
    \includegraphics[width=0.8\linewidth]{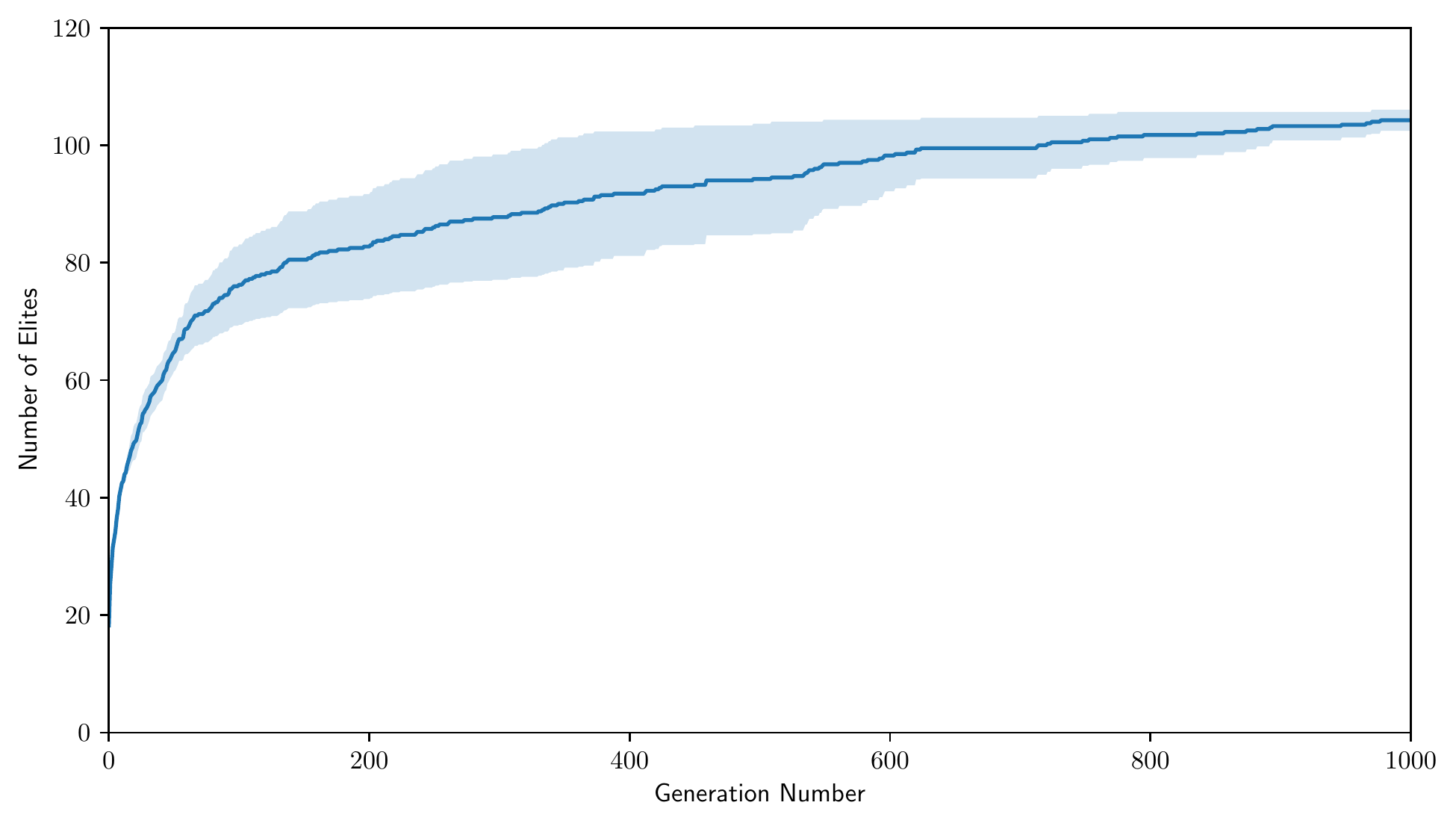}
    \caption{The average number of elites and the standard error in the Mechanics Dimensions approach over the $5$ runs.}
    \vspace{-5pt}
    \label{fig:cme_graph}
    \vspace{-10pt}
\end{figure}

The ``Mechanics Dimensions'' approach used a constrained map elites algorithm to find scenes containing the targeted mechanics. Figure~\ref{fig:cme_graph} shows the average number of elites found during $1000$ iterations. These numbers always increase, meaning that the algorithm can find more scenes as iterations pass.

\begin{figure*}
    \begin{subfigure}[t]{.28\linewidth}
        \centering
        \includegraphics[width=0.95\linewidth]{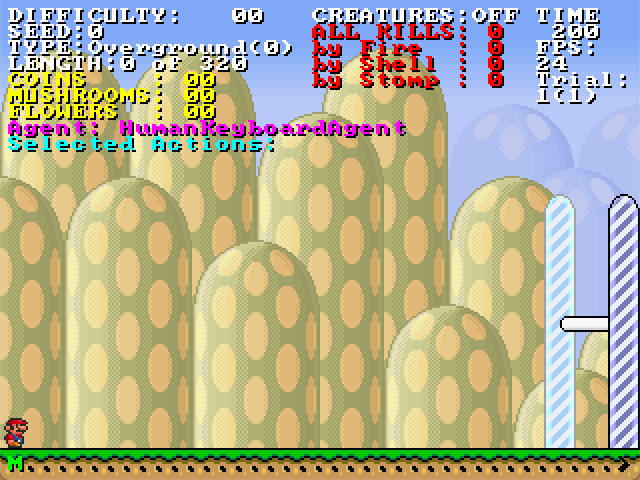}
        \caption{No fired mechancis}
        \label{fig:cme_mix_a}
    \end{subfigure}
    \begin{subfigure}[t]{.28\linewidth}
        \centering
        \includegraphics[width=0.95\linewidth]{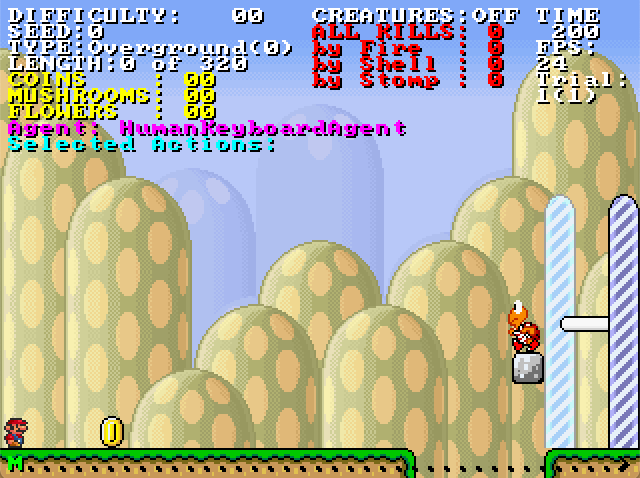}
        \caption{50\% of fired mechanics}
        \label{fig:cme_mix_b}
    \end{subfigure}
    \begin{subfigure}[t]{.28\linewidth}
        \centering
        \includegraphics[width=0.95\linewidth]{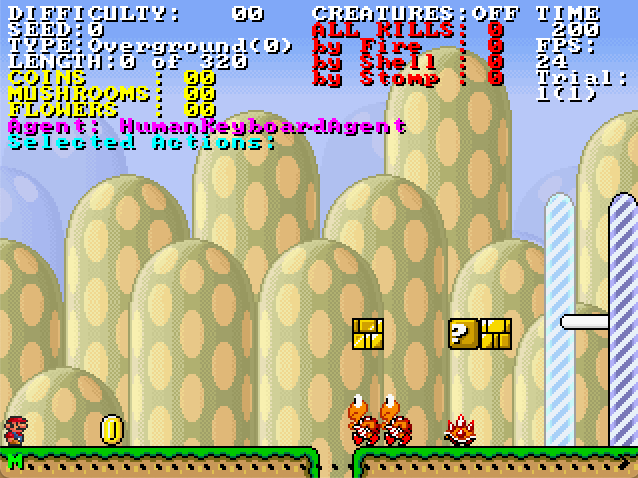}
        \caption{All fired mechanics}
        \label{fig:cme_mix_c}
    \end{subfigure}
    \vspace{-5pt}
    \caption{Three generated scenes with various degrees of number of fired mechanics.}
    \label{fig:cme_mixtures}
    \vspace{-10pt}
\end{figure*}

The constrained map-elites algorithm was able to find a huge set of different combinations of mechanics that range from having all mechanics being triggered to none of them. 
Figure~\ref{fig:cme_mixtures} shows several generated scenes with various fired mechanics. Figure~\ref{fig:cme_mix_a} doesn't need any mechanic to fire, as the floor is flat without game objects (i.e. enemies, coins, etc). Figure~\ref{fig:cme_mix_b} shows a scene that fires four different mechanics (jump, high jump, stomp, and coin) and figure~\ref{fig:cme_mix_c} shows a scene that fires all eight different mechanics. It has two enemies to guarantee shell kill, a question mark block for the mushroom, etc. However, one may notice that these scenes do not guarantee a mechanic will be fired. For example, Figure~\ref{fig:cme_mix_c} could be passed from the top of the blocks without interacting with enemies at all.

\begin{figure*}
    \begin{subfigure}[t]{.28\linewidth}
        \centering
        \includegraphics[width=0.95\linewidth]{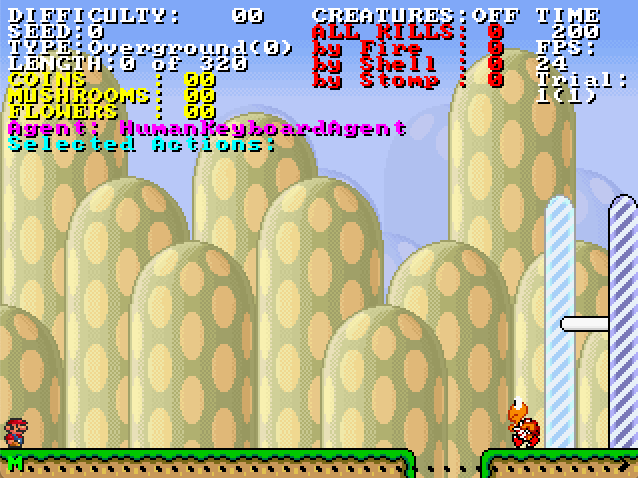}
        \caption{Enemy stomping}
        \label{fig:cme_kills_a}
    \end{subfigure}
    \begin{subfigure}[t]{.28\linewidth}
        \centering
        \includegraphics[width=0.95\linewidth]{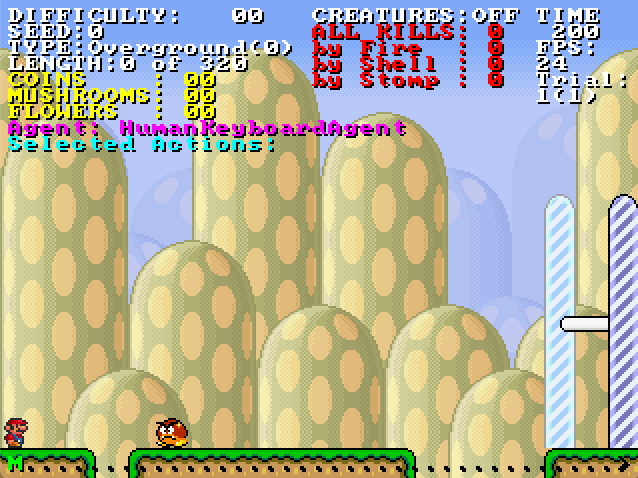}
        \caption{Enemy stomping and fall kill}
        \label{fig:cme_kills_b}
    \end{subfigure}
    \begin{subfigure}[t]{.28\linewidth}
        \centering
        \includegraphics[width=0.95\linewidth]{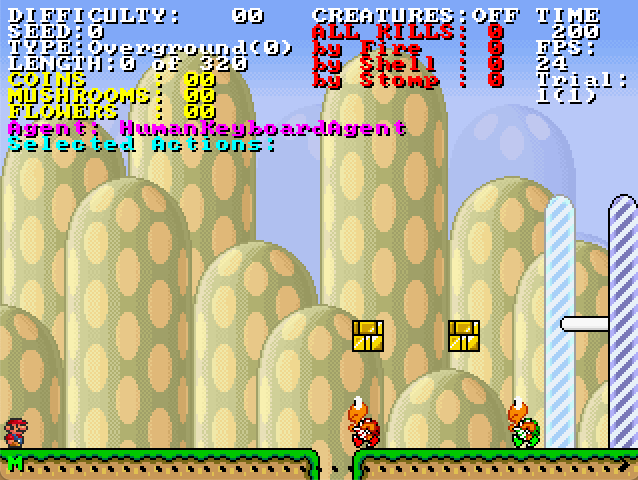}
        \caption{Enemy stomping, fall kill, and shell kill together}
        \label{fig:cme_kills_c}
    \end{subfigure}
    \vspace{-5pt}
    \caption{Three generated scenes with different ways to kill the enemies.}
    \label{fig:cme_kills}
    \vspace{-10pt}
\end{figure*}

The mechanics dimensions method contains multiple mechanics centered around how enemies can die. Figure~\ref{fig:cme_kills} shows three scenes with different types of enemy killing. For example, in a fall kill scene, the generated scene contains goombas, as they can fall of the edge (Figure~\ref{fig:cme_kills_a}). In a stomp kill scene, the generated scene contains a red koopa troopa since they cannot fall of the edge of the screen (Figure~\ref{fig:cme_kills_b}). Lastly, in scenes which target every type of kill, the generated scene contains at least two enemies, where one of them can fall of the edge (green koopa troopa) (Figure~\ref{fig:cme_kills_c}).

\begin{figure*}
    \begin{subfigure}[t]{.28\linewidth}
        \centering
        \includegraphics[width=0.95\linewidth]{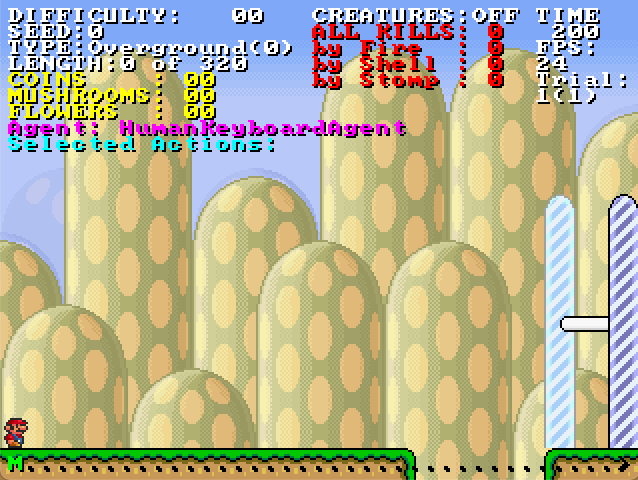}
        \caption{Normal Jump.}
        \label{fig:cme_jumps_a}
    \end{subfigure}
    \begin{subfigure}[t]{.28\linewidth}
        \centering
        \includegraphics[width=0.95\linewidth]{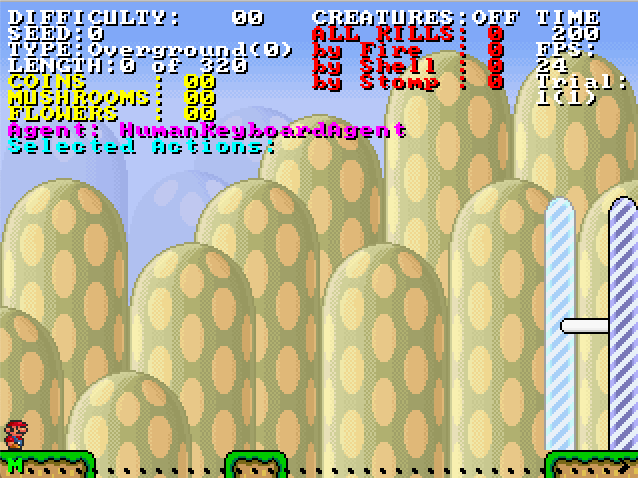}
        \caption{Long Jump}
        \label{fig:cme_jumps_b}
    \end{subfigure}
    \begin{subfigure}[t]{.28\linewidth}
        \centering
        \includegraphics[width=0.95\linewidth]{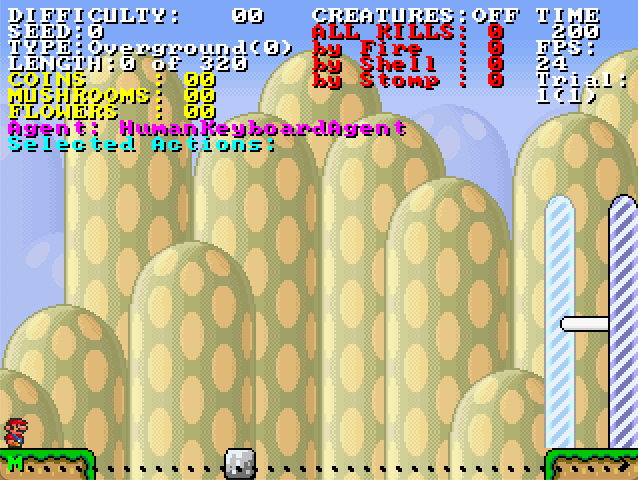}
        \caption{Long High Jump}
        \label{fig:cme_jumps_c}
    \end{subfigure}
    \vspace{-5pt}
    \caption{Three generated scenes with different ways to jump.}
    \label{fig:cme_jumps}
    \vspace{-10pt}
\end{figure*}

Similar to the kill mechanics, the mechanics dimensions method also contains a variety of jump mechanics. Figure~\ref{fig:cme_jumps} shows three  scenes targeting different ways to jump. The first scene targets the normal jump, which does not require holding the jump button for long nor running. Resulting scenes contain a small enough gap to cross (Figure~\ref{fig:cme_jumps_a}). The long jump, on the other hand, requires traversing long distances on the x-axis without the need of reaching a higher point on the y-axis (thus not requiring to hold the jump button for long). The scene in Figure~\ref{fig:cme_jumps_b} ensures that the second jump is long enough for that x-axis traversal. Lastly, the long high jump requires a longer traversing distance, i.e. pressing the jump button longer (Figure~\ref{fig:cme_jumps_c}) with a larger second gap. Similar to the punishing model experiments, all the generated level from the $5$ runs have different layout while still maintaining the necessary set-up for the mechanic to occur.

\section{Discussion}

From the results, we can see that each approach has its own advantages and disadvantages. The ``Limited Agent'' is the simpler and faster than the ``Punishing Model'', as the former doesn't involve modifying the game engine itself but only the agents. Its main disadvantage is that in order to use this technique, it is necessary to modify the agents in a way that revolves around a certain game mechanic. Modifying an agent like this can be challenging because of how difficult it can be to identify all game mechanics just from the ways the agent interacts with the world. For example, enemy blindness is easy because Mario dies when he directly collides with an enemy and therefore can be ``blind'' to the collision, but identifying a shell kill is harder (how does an agent \textit{see} or \textit{not see} a shell killing an enemy?).
Another problem is that generated scenes do not require that the player uses that mechanic in order to beat the scene. Figure~\ref{fig:la_a} shows a \textit{No Run} agent scene that the player can beat if they only wait until the koopa paratroopa passes by.

The ``Punishing Model'' was able to generate scenes that revolved around the targeted mechanic and were quite difficult to beat without triggering it. Theoretically, all the generated scenes should guarantee the required mechanic is triggered to beat it, but the usage of an imperfect agent prevented that. Because Robin Baumgarten's A* agent makes mistakes, some generated scenes were winnable even if you ignored the mechanic (e.g. Figure~\ref{fig:pm_d} and \ref{fig:pm_e}).
In most of the runs for complicated mechanics such as mushroom-related ones, the algorithm spent most generations attempting to stabilize and therefore did not have enough time to simplify its creations, resulting in complex scenes.

One can see that both FI-2Pop techniques are vulnerable to agent errors. For example, if an agent died because of a mistake the algorithm didn't intended for, the scene will pass as a good scene that contains a targeted mechanic, even if the scene might not have the mechanic anywhere in it. Possible solutions for this might be to use either more exhaustive agents, or to run the same agent multiple times and pick the best run. The trade-off with either solution is both will require more computational time.

The ``Mechanics Dimensions'' technique also does not guarantee that the player must trigger a specific mechanic to beat the scene, but it does guarantee that the mechanic \textit{could} be used if the player desired. It was able to generate scenes with more than one targeted mechanic at the same time, and it used a single agent in the generation process, making it the fastest approach of all three (taking half the required time of the others). Furthermore, the ``Mechanics Dimensions'' method does not search for one scene like the other two approaches, instead searching for a set of scenes that contain the targeted mechanic(s). This way of thinking might not be optimal in some cases, e.g. if the target mechanic is not that easy to find, the algorithm might spend too much time to find other unnecessary mechanics. To modify this technique to guarantee that it generates scenes in which a targeted mechanic must occur, the playing agent could theoretically be swapped for an agent that is able to find \textit{all possible solutions} in a scene. Then the target mechanic is recorded as fired only if it is triggered in all possible solutions. This would however be impractical even for a game of Super Mario Bros' complexity.

\section{Conclusion}

An important problem within game AI is to generate levels with specific characteristics for given domains. Game levels are often used to teach mechanics, or to at least guarantee that players are skilled enough at using a certain mechanic in order to advance in the game. This paper introduced three methods for automatically creating levels that revolve around specific mechanics. We used the Mario AI framework as a testbed and focused on generating scenes, small sections of a level that tackle a certain idea or mechanic(s). 

While the results shown are promising, there are still limitations. The first approach, ``Limited Agents'', relies on manually identifying mechanics and implementing agents that have limited knowledge of each one. Both this and the second approach, ``Punishing Models'', are susceptible to agent-related game-playing mistakes. Also, both the first and the third approach, ``Mechanics Dimensions'', don't guarantee that the target mechanics is needed to beat the scene. In the future, we may address these limitations by building more robust agents, so we can exhaustively search possible solutions. The generated levels are also only scenes. By connecting these scenes together, our approaches might be able to construct entire levels.

This work can also be of benefit to and benefit from works that focus on extracting mechanics from games, such as that at \textit{AtDelfi Project}~\cite{green2018atdelfi}, which identifies mechanics from a critical path to win the game. We could use these to automatically identify dimensions in the Constrained Map-Elites.
    

\appendix

\begin{acks}
Ahmed Khalifa acknowledges the financial support from NSF grant (Award number 1717324 - ``RI: Small: General Intelligence through Algorithm Invention and Selection.''). Michael Cerny Green acknowledges the financial support of the SOE Fellowship from NYU Tandon School of Engineering.
\end{acks}

\bibliographystyle{ACM-Reference-Format}

\end{document}